\pdfoutput=1

\documentclass[11pt]{article}

\usepackage[final]{acl}

\usepackage{times}
\usepackage{latexsym}

\usepackage[T1]{fontenc}

\usepackage[utf8]{inputenc}

\usepackage{microtype}

\usepackage{inconsolata}

\usepackage{xspace}
\usepackage{xcolor}
\usepackage{amssymb}
\usepackage{multirow}
\usepackage{floatrow}
\usepackage{soul}
\usepackage{url}
\usepackage[utf8]{inputenc}
\usepackage{graphicx}
\usepackage{amsmath}
\usepackage{amsthm}
\usepackage{booktabs}
\usepackage{algorithm}
\usepackage{algorithmic}
\usepackage{graphicx}
\usepackage[group-separator={,}]{siunitx}
\usepackage{subcaption}

\usepackage{comment}
\usepackage[inline]{enumitem}

\DeclareMathOperator*{\argmax}{argmax}

\title{Word Sense Linking: Disambiguating Outside the Sandbox}

\author{Andrei Stefan Bejgu$^{1,2}$, Edoardo Barba$^{1}$, {\bf Luigi Procopio}$^3$ \\ {\bf Alberte Fernández-Castro}$^4$ \normalfont{and} {\bf Roberto Navigli}$^1$\\
$^1$Sapienza NLP Group, Sapienza University of Rome\\
$^2$Babelscape, Italy \qquad 
$^3$Litus AI\\
$^4$Roma Tre University\\
\texttt{\{lastname\}@diag.uniroma1.it}, \texttt{bejgu@babelscape.com}
}

\begin{document}
\maketitle
\begin{abstract}
Word Sense Disambiguation (WSD) is the task of associating a word in a given context with its most suitable meaning among a set of possible candidates. While the task has recently witnessed renewed interest, with systems achieving performances above the estimated inter-annotator agreement, at the time of writing it still struggles to find downstream applications. We argue that one of the reasons behind this is the difficulty of applying WSD to plain text. Indeed, in the standard formulation, models work under the assumptions that a) all the spans to disambiguate have already been identified, and b) all the possible candidate senses of each span are provided, both of which are requirements that are far from trivial. In this work, we present a new task called Word Sense Linking (WSL) where, given an input text and a reference sense inventory, systems have to both identify which spans to disambiguate and then link them to their most suitable meaning.
We put forward a transformer-based architecture for the task and thoroughly evaluate both its performance and those of state-of-the-art WSD systems scaled to WSL, iteratively relaxing the assumptions of WSD. 
We hope that our work will foster easier integration of lexical semantics into downstream applications.

\end{abstract}

\section{Introduction}
\label{sec:intro}

Leveraging the advances in pretrained transformer architectures \cite{devlin-etal-2019-bert}, Word Sense Disambiguation (WSD) systems have nowadays reached performances on several evaluation benchmarks that are on par with their estimated inter-annotator agreement \cite{bevilacqua-navigli-2020-breaking, barba-etal-2021-consec}.
However, despite these advances, the task is still well known for struggling to find downstream applications.
We argue that one of the possible causes is the difficulty of applying WSD in unconstrained settings due to its heavy assumptions. Indeed, three requirements need to be met for a generic state-of-the-art system to perform WSD on some input text:
\begin{enumerate*}[label=\roman*)]
    \item[R1)] a sense inventory, that is, a semantic resource providing a comprehensive list of all the senses of interest, must be provided,
    \item[R2)] the list of spans to disambiguate in the input text must have already been identified, and
    \item[R3)] an oracle that pairs each span to its set of possible senses, realized through a manually curated word-to-sense mapping, must be available.
\end{enumerate*} 

While the first condition is intrinsic to the disambiguation objective and, thus, unavoidable, we argue that the second two (i.e. R2 and R3) are system-specific assumptions that can be relaxed.
We formulate this idea into a new task called Word Sense Linking (WSL), which more accurately reflects the conditions of downstream applications such as Neural Machine Translation \cite{liu-etal-2018-handling, iyer-etal-2023-code}, Information Extraction \cite{moro2013integrating, delli-bovi-etal-2015-large} and the enrichment of contextual representations in neural models \cite{peters-etal-2019-knowledge}. 
In WSL, given an input text and a reference sense inventory, systems have to identify which spans to disambiguate and link them to their most suitable meaning in the sense inventory.

Similarly to Entity Linking \cite{broscheit-2019-investigating}, WSL can be split into three simpler subtasks, with traditional WSD taking place after two initial stages of Concept Detection (CD), that is, the identification of the spans to be disambiguated in the input text (R2), and Candidate Generation (CG), the generation of a list of sense candidates for each target span (R3).
For example, given some reference sense inventory and the sentence \textit{``Bus drivers drive buses for a living.''}, Concept Detection might identify \texttt{[bus drivers, drive, buses, living]} as the spans to disambiguate, while Candidate Generation might provide \texttt{[vehicle, electrical conductor]} as the sense candidates for \textit{buses}.

In this work, we first formally introduce the task of Word Sense Linking, along with its three components, and, then, put forward a novel architecture for the task, based on the retriever-reader paradigm \cite{karpukhin-etal-2020-dense}. We thoroughly study the behavior of our architecture and that of state-of-the-art WSD systems, once they have been scaled to WSL, as we iteratively relax the sandboxed assumptions of WSD, starting with R2 and continuing with R3. Our analysis highlights a number of important -- yet neglected -- challenges when it comes to performing disambiguation in unconstrained settings. In particular, straightforward and natural extensions of WSD systems to WSL result in large performance drops. In contrast, our model demonstrates considerably more robustness and consistently outperforms such extensions of WSD systems by a large margin. The contributions of this work are therefore as follows:
\begin{itemize}
    \item We introduce the task of Word Sense Linking, which we believe to better represent the settings of downstream applications for WSD.
    \item We introduce for the first time a Word Sense Linking evaluation dataset, enriching the \textit{de-facto} standard WSD benchmark \cite{raganato-etal-2017-word}.
    \item We put forward a novel flexible architecture for this task and evaluate, in multiple settings, both its behavior and that of state-of-the-art WSD systems scaled to WSL.
    \item Overall, our findings underline several crucial yet neglected challenges when scaling WSD systems to a real-world scenario.
\end{itemize}

We release code, data, and model weights at \url{https://github.com/Babelscape/WSL}.

\section{Word Sense Linking}
Word Sense Linking is the task of identifying and disambiguating all the spans of an input text with their most suitable senses chosen from a reference inventory. Formally, let $t$ be the input text, with $t_1, \dots, t
_{|t|}$ being its words, and $I$ the reference inventory, containing a set of senses. 
Then, a WSL system can be represented as a function $f$ that takes as input the tuple $(t, I)$ and outputs a list of triples $[(s_1, e_1, g_1), \dots, (s_n, e_n, g_n)]$ where each triple $(s_i, e_i, g_i)$, $i \in [1,n]$, represents a disambiguated span, with $s_i$ and $e_i$ being the start and end token index of the span, and $g_i \in I$ representing the corresponding sense chosen from the inventory. Conceptually, WSL can be divided into three subtasks, namely, Concept Detection, Candidate Generation, and Word Sense Disambiguation.

\paragraph{Concept Detection (CD)} is the task of identifying the spans to disambiguate in an input text $t$ given a reference inventory $I$.
A Concept Detection system can be represented as a function that takes as input the tuple $(t, I)$ and outputs a list of pairs $[(s_1, e_1), \dots, (s_n, e_n)]$, each marking the boundaries of a span to disambiguate. Specifically, $\forall i \in [1, n]$,  $s_i$ and $e_i$ are the start and end token index of the $i$-th span to disambiguate.

\paragraph{Candidate Generation (CG)} is the task of generating a set of possible candidate senses that an input span occurring in some text $t$ can assume given the reference inventory $I$.
A Candidate Generation system can be represented as a function that takes as input $t$, $I$ and the start and end token indices $(s, e)$ of the span, and outputs a set of sense candidates $PC \subseteq I$,  that the span can assume.

\paragraph{Word Sense Disambiguation (WSD)} aims at identifying, for each target span occurring in an input text $t$, the most suitable sense among a set of possible candidates.
A WSD system can be represented as a function that takes as input $t$, $I$, a list of span indices $[(s_1, e_1), \dots, (s_n, e_n)]$ and corresponding sets of possible candidates $[PC_1, \dots, PC_n]$. It then outputs a list of triples $[(s_1, e_1, g_1), \dots, (s_n, e_n, g_n)]$ where, $\forall i \in [1, n]$, each span $(s_i, e_i)$ is paired with the sense $g_i$ chosen by the system among the candidates provided in $PC_i$.

\paragraph{Finally} it is easy to see that a WSL system can be built by concatenating, in cascade, a Concept Detection, a Candidate Generation, and a Word Sense Disambiguation module. 
This structure not only facilitates the extension of WSD systems to WSL but also serves as a flexible framework rather than a strict recipe. In fact, as we will show in the next section, the architecture we put forward does not follow this flow and inverts the CD and CG steps.

\begin{figure*}[h!]
    \centering
    \includegraphics[width=0.60\textwidth, trim={0 2.0cm 0 2.1cm},clip]{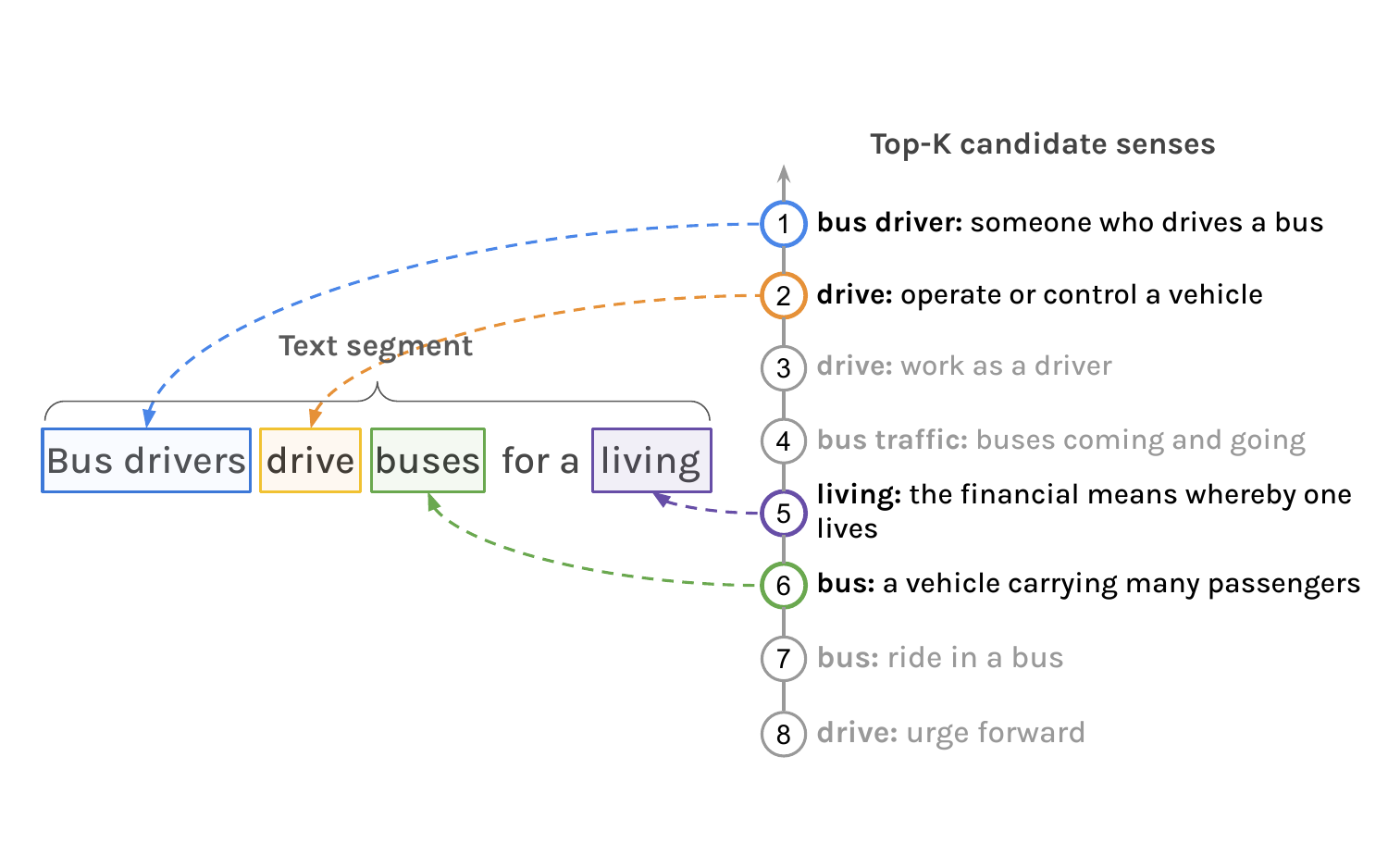}
    \caption{Our WSL process. First, the retriever identifies the top-$k$ candidate senses (Candidate Generation). Then, the reader identifies the spans to be disambiguated (Concept Detection) and pairs each of these with their most suitable sense (Word Sense Disambiguation).}
    \label{fig:architecture}
\end{figure*}

\section{Model}

The flow \textit{CD-then-CG} presents an intrinsic limitation: each span identified by Concept Detection ought to correspond to the occurrence of a specific sense candidate among those produced by Candidate Generation and, yet, CG occurs after CD. 
That is, we need to identify the span we will, later on, link to a specific sense without actually knowing this sense.
To overcome this limitation, inspired by the work presented by \citet{zhang-etal-2022-entqa} for Entity Linking, we invert the CD and CG steps, and propose a novel flexible architecture for WSL based on the retriever-reader paradigm. In this section, we first outline its formulation (Section~\ref{sec:model-formulation}) and, then, discuss its two components, namely the retriever (Section~\ref{sec:model-retriever}) and the reader (Section~\ref{sec:model-reader}).

\subsection{Formulation}
\label{sec:model-formulation}

Starting from the input text $t$ and the reference inventory $I$, our formulation is as follows. First, we perform CG on the entire input text, producing an ordered list of unique candidates $PC(t) \subseteq I$, each coming from $I$ and likely to represent the meaning of some arbitrary span in $t$. Then, moving to CD and using vector representations contextualized on both $t$ and $PC(t)$, we let a classifier identify the start and end token indices of every span in $t$; this operation results in a list of $n$ tuples $[(s_1, e_1), \dots, (s_n, e_n)]$. Finally, we perform WSD on the identified spans, pairing each $(s, e)$ with its most suitable sense $g^{*} \in PC(t)$.\footnote{We extend the $\in$ operator to work on lists as well.}
Figure \ref{fig:architecture} reports a visual outline of this process.

We implement our formulation with transformer-based architectures that operate on textual inputs. However, this makes it necessary for our inputs to have such representations and, while $t$ inherently satisfies this requirement, the same does not hold for the senses in $I$. To overcome this issue, for each sense $g \in I$, we define a textual representation, which we build by concatenating its lemmas\footnote{If a mapping between senses and words is available.}, provided through the mapping from word to possible senses, and its textual definition in $I$.\footnote{Although we did not explicitly mention it before, it is customary for sense inventories to provide, for each of their senses, a textual definition (\textit{gloss}) that defines its meaning.}

Finally, we note that $t$ can be arbitrarily long. Consequently, providing the entire $t$ as input to our formulation can be challenging, both computationally and from a modeling perspective. To overcome this, we adopt a sliding window approach, with window size $w$ and stride $\tau,$ and actually feed, to the previous steps, single chunks of $t$, rather than the entire sequence. Once all chunks have been processed for $t$, we retain only \textit{conflict-free predictions}, i.e., every triple $(s, e, g^{*})$ such that the processing of every chunk that includes $(s, e)$ in its window always results in assigning the same  $g^{*}$ to $(s, e)$. Henceforth, with no loss of generality, we replace our input, and the meaning of notation $t$, with a generic chunk, rather than the entire sequence.

\subsection{Retriever}
\label{sec:model-retriever}

We implement the initial Candidate Generation step via dense passage retrieval \cite{karpukhin-etal-2020-dense}, using a transformer-based retriever consisting of an encoder $E$ to produce a dense representation of text passages and senses.
Starting from the input text $t$ and the inventory $I$, we use $E$ to compute a vector representation $v_t$ for $t$, and $v_g$ for every sense $g \in I$. Then, we use the dot product $v_t \cdot v_g$ to rank all the senses in $I$ and, finally, extract the top $k$ among these. 
The resulting $g^1, \dots, g^k$ senses constitute our sense candidate set $PC(t)$ for $t$.

To train our model, we use a multi-label variant of noise contrastive estimation \cite[NCE]{zhang-etal-2022-entqa}, maximizing the following objective:
\begin{align*}
    \sum_{g \in \Gamma(t)} \log \frac
        {\exp (v_t \cdot v_g)}
        {\exp (v_t \cdot v_g) + \sum_{g^{\prime} \in \mathrm{N(t)}} \exp (v_t \cdot v_{g^{\prime}})}
\end{align*}
where $\Gamma(t) \subset I$ is the set of gold senses appearing in $t$ and $N(t) \subset I \setminus \Gamma(t)$ represents a selection of negative examples. Considering the crucial role of $N(t)$ in retriever-based architectures \cite{karpukhin-etal-2020-dense}, we adopt a strategy aimed at selecting adequately informative negative examples, while, at the same time, accounting for the presence of unannotated tokens. 
Starting from $I\setminus \Gamma(t)$, we first discard the senses that never appear in the training set or whose part of speech does not match any sense-annotated span in $t$. 

Then, denoting by $I^R(t)$ the set of resulting senses in $I$ for $t$, we define $N(t)$ as the union of the two sets $N_1(t)$, a collection of hard negatives \cite{gillick-etal-2019-learning} in $I^R(t)$, and $N_2(t)$, a subsample of $I^R(t)$ that aims at counterbalancing the bias towards the most frequent senses in the training corpus.
Specifically, we build $N_1(t)$ by selecting the $\nu_1$ senses in $I^R(t)$ to which the retriever assigns the highest score, and $N_2(t)$ constructed using $\nu_2$ gold senses from other samples in the same mini-batch.

\subsection{Reader}
\label{sec:model-reader}

Having identified the sense candidates $PC(t) = g^1, \dots, g^k$, we now describe the remaining Concept Detection and WSD steps, which we formulate as a multi-task multi-label classification problem. To this end, we concatenate $t$ and $g^1, \dots, g^k$ into a single sequence $m$, prepending a special symbol $[S_i]$  for each sense candidate $g^i$:
\begin{align}
    m = t\ [S_1]\ g^1\ \dots\ [S_k]\ g^k
    \label{eq:m}
\end{align}
We feed $m$ to a transformer encoder, producing a series of vector representations for the tokens in $m$; let $h_1, \dots, h_{|t|}$ be the representations corresponding to the tokens in $t$, and $h_g$ the one associated with the special symbol of a generic sense candidate $g \in PC(t)$. With these contextualized representations, we realize CD and WSD as follows.

We begin with CD, identifying the spans to disambiguate. Specifically, we apply two classification heads, namely $H_{start}$ and $H_{end}$, on each representation to determine whether the corresponding token is a start or end of some span in $t$; both heads consist of two linear transformations with a ReLU activation in between. However, their behavior -- and input -- differ: while $H_{start}$ receives single token representations, $H_{end}$ operates at span level. That is, once $H_{start}$ has identified some index $s$ as a span start, to determine whether some other index $e \geq s$ is its end, $H_{end}$ takes as input $[h_s; h_e]$, that is, the concatenation of $h_s$ and $h_e$. Once completed, this step results in a list of span indices $[(s_1, e_1), \dots, (s_n, e_n)]$.

Then, moving to WSD, we let each extracted span $(s, e)$ identify the sense $g^{*} \in PC(t)$ it refers to. We start by computing the following vectors:
\begin{align*}
    h_{se}^{'} &= [f_1(h_s); f_2(h_e)] \\
    h_{g}^{'} &= [f_1(h_g); f_2(h_g)] \; \; \forall g \in PC(t)
\end{align*}
where $f_1$ and $f_2$ are both functions consisting of two identical but independent linear transformations, interleaved by a ReLU activation. Then, we pair the span $(s, e)$ with the sense $g^{*} = \argmax_{g \in PC(t)} \; h_{se}^{'} \cdot h_g^{'}$.
We replicate this strategy for all the extracted spans, hence eventually producing as output $[(s_1, e_1, g_1^{*}), \dots, (s_n, e_n, g_n^{*})]$. Our reader is trained by jointly maximizing three cross-entropy objectives, respectively over 
\begin{enumerate*}[label=\roman*)]
    \item the gold start indices in $t$,
    \item the gold end indices in $t$ and
    \item the gold sense for every span in $t$.
\end{enumerate*}

We note that our architecture presents a number of interesting properties. 
First, since $h_1, \dots, h_{|t|}$ are vectors contextualized over the entire input sequence $m$ (Equation \ref{eq:m}), which includes $g^1, \dots, g^k$, the choice of the extracted spans does indeed take into account the sense candidates available. Second, this contextualization is not limited to $h_1, \dots, h_{|t|}$ but, in fact, applies to $h_{g^1}, \dots, h_{g^k}$ as well. This means that the sense candidates are also contextualized on each other, allowing for better representations, as in \citet{barba-etal-2021-esc}. Third, computationally speaking, the only demanding operation corresponds to the encoding of $h_1, \dots, h_{|t|}$. However, since this operation depends only on $t$ and $PC(t)$, we can disambiguate all the spans in $t$ in a single forward pass, which makes it very fast and hence suited for downstream applications.

\section{Evaluation}

We now investigate the effectiveness of our model and how it relates to the R2 and R3 assumptions of WSD (Section~\ref{sec:intro}). 
To this end, we first outline its implementation details (Section \ref{sec:evaluation-model}) and the novel dataset introduced for the WSD and WSL evaluation (Section \ref{sec:wsl-benchmark}), which will remain unchanged throughout our experiments. Then, we evaluate it on plain English WSD (Section~\ref{sec:evaluation-wsd}), so as to set an initial reference for what regards its disambiguation capabilities. Finally, we move to WSL and gradually relax these sandboxed assumptions, dropping R2 (Section~\ref{sec:evaluation-wsd-cd}) and examining different relaxations on R3 (Section~\ref{sec:evaluation-wsd-cd-cg}).

\subsection{Model Details}
\label{sec:evaluation-model}

\paragraph{Formulation Hyperparameters} We use $w=16$ and $\tau=8$ for our sliding window approach. 
This ensures that each window captures sufficient context for an effective disambiguation while maintaining a manageable overlap between consecutive windows.

\paragraph{Retriever} We employ mean pooling over $E5_{base}$ \cite{wang2022text}, a pretrained transformer-based architecture, for our retriever encoder $E$. The entire system is trained with a batch size of \num{16} input texts for \num{150000} steps using AdamW \cite{loshchilov2018decoupled}, with \num{20}\% warm-up and $2 \cdot 10^{-5}$ learning rate, and setting $k = 100$. At training time, we construct the sense candidate set by including the gold senses, $\nu_1=5$\% of negatives from $N_1(t)$ and the remaining $\nu_2$ from $N_2(t)$. 

\paragraph{Reader} We use the base version of DeBERTa-v3 \cite{he2021debertav3} as the transformer-based encoder in our reader. The four linear transformations, namely two for span detection and two for Word Sense Disambiguation, present the same intermediate dimensionality, that is, \num{768}; instead, the final dimension is \num{1} for the span detection mappings, as they are classification heads, but \num{768} for the WSD mappings. The overall system is trained with a batch size of \num{4096} tokens for \num{100000} steps using AdamW \cite{loshchilov2018decoupled} with a learning rate of $10^{-4}$ and a layerwise decay rate of \num{0.8}.

\subsection{WSL Benchmark}
\label{sec:wsl-benchmark}
The framework presented by \citet{raganato-etal-2017-word} represents the de facto standard benchmark for WSD and is based on the WordNet sense inventory \cite{milleretal:90}. 
Specifically, it is composed of Senseval-2 \cite[\textbf{SE02}]{edmonds-cotton-2001-senseval}, Senseval-3 \cite[\textbf{SE03}]{snyder-palmer-2004-english},  SemEval-2007 \cite[\textbf{SE07}]{pradhan-etal-2007-semeval}, SemEval-2013 \cite[\textbf{SE13}]{navigli-etal-2013-semeval} and SemEval-2015 \cite[\textbf{SE15}]{moro-navigli-2015-semeval}. 
The original task datasets might include spans of text that contain content words but were not assigned any specific meaning by annotators from among those contained in the reference sense inventory.  
This could have occurred for a variety of reasons, such as in the case of SE13, where only nouns were chosen to be annotated.
Whereas for the WSD setting the absence of this information does not pose a problem for evaluation, for WSL it renders evaluation impossible.
Specifically, we cannot in such case measure the precision of a WSL system as we cannot discern between wrongly predicted spans and annotation omissions. 
To address this issue and overcome the lack of a WSL-specific dataset, we introduce a dedicated evaluation resource aimed at bridging the annotation gap, not just in terms of precision, but also in terms of recall. 
We comprehensively annotated the standard WSD evaluation datasets, increasing the annotation count from $7253$ to $11623$ annotations, and resulting in the complete coverage of all the content words. 

This substantial increase in annotations allows for a comprehensive benchmark, facilitating future research in the field by providing a more robust framework for evaluating the precision, recall, and F1 of WSL systems.

\paragraph{Annotation process}
\label{sec:wsl-benchmark-annotation}

We have annotated the standard WSD evaluation dataset (i.e., ALL, the one utilized in the previously presented WSD evaluation framework \cite{raganato-etal-2017-word}). This process aimed to ensure a comprehensive and accurate representation of terms and their meanings, using WordNet as the sense inventory.
We selected these guidelines for the annotators:
\begin{enumerate}
    \item \textbf{Multiwords:} Annotators marked terms such as "lung cancer" as single entities, focusing on those recognized in WordNet with contextually coherent meanings.  
    \item \textbf{Sub-words:} Annotators also marked the individual components of multiword expressions, but only when the sub-words had coherent meanings within the sentence context. This dual-level annotation strategy captured both the general and specific meanings of terms.
    \item \textbf{Non-content Words}: Annotators excluded non-content words, such as auxiliary verbs, from annotation. These words are essential for grammatical structure but do not carry any semantic weight.
\end{enumerate}

Employing WordNet as the sense inventory facilitated a uniform and precise approach to annotation across the entire dataset and maintained consistency with WSD tasks, ensuring alignment with established standards and methodologies in the field.

Due to the complexity of the task, a single expert linguist, who is also an author of this paper and has a robust background in the annotation of lexical-semantic tasks, conducted the majority of the annotation work. Nonetheless, to validate the reliability and consistency of the annotation process, we conducted an inter-annotator agreement evaluation. This involved a subset of the dataset, constituted by 20 sentences independently annotated by three different expert linguists,\footnote{We paid the annotators according to the standard salary for their geographical location.} resulting in 222 distinct annotations. The agreement among annotators was measured using Cohen’s kappa statistic, which yielded a score of 0.83, interpreted as an almost perfect level of agreement, showing a high degree of annotation consistency.
These results underscore the robustness and reliability of our annotation methodology, despite the inherent challenges of subjective interpretation and the detailed demands of manual annotation. 
Further statistics about the newly introduced dataset can be found in \autoref{sec:appendix_ann}. 

\subsection{Word Sense Disambiguation}
\label{sec:evaluation-wsd}

\paragraph{Setting} We evaluate our model on all-word English WSD using WordNet as our sense inventory. 
WordNet provides a comprehensive and structured database of English word senses, making it a widely accepted benchmark for WSD. By utilizing WordNet, we ensure that our evaluation aligns with established standards in the field, facilitating comparison with previous work and other state-of-the-art models.

\paragraph{Comparison Systems} We compare our model with recent state-of-the-art systems for WSD, which we divide into two different categories. On the one hand, we consider systems that frame WSD as a sequence-level classification problem, that is, systems that disambiguate a single span at a time. We include in our evaluation: \citet[\textbf{ESCHER}]{barba-etal-2021-esc}, the first approach reformulating WSD as a text extraction problem; \citet[\textbf{KELESC}]{zhang-etal-2022-word}, a knowledge-enhanced version of ESCHER, incorporating additional information coming from WordNet; \citet[\textbf{ESR}]{song-etal-2021-improved-word}, an architecture framing WSD as a binary classification problem; and \citet[\textbf{ConSeC}]{barba-etal-2021-consec}, the system that, thanks to iterative disambiguation, holds the state of the art on the reference benchmark at the time of writing. 

On the other hand, we report models that frame WSD as a token-level classification problem, thus disambiguating all the spans in a sentence together. We include: \citet[\textbf{BEM}]{blevins-zettlemoyer-2020-moving}, a bi-encoder system incorporating gloss knowledge; \citet[\textbf{EWISER}]{bevilacqua-navigli-2020-breaking}, a classifier modeling the relational knowledge in WordNet using a Personalized PageRank approach; and \citet[\textbf{WMLC}]{conia-navigli-2021-framing}, a classification formulation for WSD that leverages the relational knowledge in WordNet at training time. We use this division to highlight not only the different designs of our comparison systems but, also, their corresponding trade-offs. Indeed, while sequence-level classifiers typically achieve higher performances, token-level classifiers emphasize speed and are considerably more usable in downstream applications.

\paragraph{Data} We evaluate our model using the framework presented by \citet{raganato-etal-2017-word}. 
Specifically, we train our system on SemCor \cite{Milleretal:93}, perform model selection on SE07, and test on the other available datasets. 
We measure performances in terms of F1 score and, as in previous works, report this score on the concatenation of each evaluation dataset (identified as \textbf{ALL}).
Finally, to set a reference for the WSL setting, we evaluate our model and the comparison systems on the novel benchmark presented in Section \ref{sec:wsl-benchmark} under the WSD assumptions (identified as \textbf{ALL}$_{FULL}$).

\paragraph{Our Model Behavior} Comparison systems operate in this setting assuming the availability of an oracle for both CD and CG. To enable a fair comparison, we 
\begin{enumerate*}[label=\roman*)]
    \item limit the candidates retrieved from the retriever module only to those that the spans to disambiguate can assume in WordNet and
    \item force the reader to select for each span only the senses that it can assume in WordNet.
\end{enumerate*}

\begin{table}[t]
    \center
    \small
    \begin{tabular}{clc|lll}
    \toprule
    & Models & Params & SE07 & ALL & ALL$_{FULL}$  \\
    \midrule
    \parbox[t]{2mm}{\multirow{4}{*}{\rotatebox[origin=c]{90}{\textit{Sequence}}}}
    & ESCHER                      &      400M &      76.3 &    80.7 & 81.2 \\
    & KELESC                                 &      400M &      76.7 &    81.2 & 81.4 \\
    & ESR                               &      350M &      77.0 &    81.1 & 81.3 \\
    & ConSeC & 400M &      \textbf{77.4} &    \textbf{82.0} & \textbf{82.5} \\
    \cmidrule(lr){1-6}
    \parbox[t]{2mm}{\multirow{4}{*}{\rotatebox[origin=c]{90}{\textit{Token}}}}
    & WMLC &  340M  &  72.2 &  77.6 & 78.1 \\
    & EWISER & 340M   & 71.0 & 78.3 & 78.9 \\
    & BEM & 220M   & 74.5 & 79.0 & 79.7 \\
    & Our Model     &      295M &      \textbf{75.2} &      \textbf{80.2} & \textbf{80.8}\\
    \bottomrule
    \end{tabular}
    \caption{WSD results for sequence-level and token-level classifiers.}
    \label{tab:wsd-results}
\end{table}

\paragraph{Results}

Table \ref{tab:wsd-results} shows how all the systems under comparison fare on the standard framework. As a first result, we note that our model, a token-level classifier as all spans are disambiguated together, outperforms all the other token-level systems, almost reaching sequence-level ESCHER's performance.
Furthermore, while the difference in comparison to ConSeC, i.e., the best-performing model, is significant in terms of F1 score (\num{1.8} points), the speed comparison reveals a compelling advantage: using ALL as a reference, our model can process the \num{7253} instances in less than \num{17} seconds, while ConSeC requires a total of \num{73} seconds.\footnote{To perform the comparison, we use the code made available by the authors at \url{https://github.com/SapienzaNLP/consec} using an RTX 3090 for both experiments.} 
These results suggest that our model provides a solid foundation for our setting, showing not only strong performance on WSD, but also that it is flexible enough to support CD and CG.
Finally, and most importantly, we can see that the performances obtained by the comparison systems on the \textbf{ALL}$_{FULL}$ dataset align with those on the standard benchmark, allowing us to study how they change when relaxing the WSD constraints
.

\subsection{Word Sense Linking: Dropping the Concept Detection Oracle}
\label{sec:evaluation-wsd-cd}

\paragraph{Setting} Here, we drop the R2 condition. 
In the context of WSL, this means that the system is required to identify spans of text that need to be disambiguated. 
By removing this oracle, we simulate a more challenging and practical setting where the system has to detect the spans to disambiguate before proceeding to candidate generation and disambiguation.

\paragraph{Comparison Systems} Given the task's novelty, there are no direct comparison systems. We take advantage of this opportunity to assess the complexity of Concept Detection and its impact on ConSeC and BEM, that is, the best sequence-level and token-level classifiers described in the previous section. Specifically, to establish baselines, we implement two natural straightforward solutions for CD and pipeline each of these with our reference WSD systems. Our first solution consists of a simple heuristic approach: given an input text, we find all the longest spans such that the corresponding lemma is linked, through the mapping from word to possible senses, to any sense in WordNet. As an alternative to this strategy, we consider a supervised implementation to determine whether each token represents the start, inside, or outside of a span to disambiguate. We train\footnote{We use the span information contained in SemCor for training and SemEval-2007 for model selection.} a variant of the architecture of \citet{mueller-etal-2020-sources} widely used in Named Entity Recognition tasks using BERT-base as an encoder. We denote the four variants by BEM$_{SUP}$, BEM$_{HEU}$, ConSeC$_{SUP}$ and ConSeC$_{HEU}$.

\begin{table}[t]
    \center
    \small
    \begin{tabular}{lcccccc}
    \toprule
    \multicolumn{1}{c}{} & \multicolumn{3}{c}{\textbf{SE07}} & \multicolumn{3}{c}{\textbf{ALL$_{FULL}$}} \\ 
    \cmidrule(lr){2-4} \cmidrule(lr){5-7}
    Models & P & R & F1 & P & R & F1 \\
    \midrule
    BEM$_{SUP}$ & 67.6 & 40.9 & 51.0 & 74.8 & 50.7 & 60.4 \\
    BEM$_{HEU}$ & 70.8 & 51.2 & 59.4 & 76.6 & 61.2 & 68.0 \\
    ConSeC$_{SUP}$ & 76.4 & 46.5 & 57.8 & 78.9 & 53.1 & 63.5 \\
    ConSeC$_{HEU}$ & \textbf{76.7} & 55.4 & 64.3 & \textbf{80.4} &   64.3 & 71.5 \\
    Our Model & 73.8 & \textbf{74.9} & \textbf{74.4} & 75.2 & \textbf{76.7} & \textbf{75.9} \\
    \bottomrule
    \end{tabular}
    \caption{WSL results with no CD oracle.}
    \label{tab:wsl-results}
\end{table}

\paragraph{Data} To address the issue of missing annotations in the training phase of our WSL system, specifically when using SemCor, which is known for its incomplete annotations, we propose a mitigation strategy.
We use ConSeC$_{HEU}$ to identify and annotate all the missing spans in SemCor, leading to the creation of SemCor$_C$. This procedure results in \num{133727} new annotations, in addition to the original \num{226036} already in SemCor. We note that the usage of these annotations proves to be crucial for our model but irrelevant, if not actually harmful in some cases, for both BEM and ConSeC. Further details are available in \autoref{sec:app_semcor}. Therefore, in what follows, the results we report always refer to the usage of SemCor$_C$ for our model and of solely SemCor for our comparison systems.

\paragraph{Our Model Behavior} Differently from the previous section, here we run the model in a WSL setting. The retriever identifies the sense candidates and the reader their corresponding spans. However, as the other systems use a CG oracle, for a fair comparison, for a given identified span, we limit the reader to selecting only the senses it can assume.

\paragraph{Results}

\begin{table}[t]
    \center
    \small
    \begin{tabular}{lc|cccc}
    \toprule
    \cmidrule(lr){3-5}
    Models & Lemmas & P & R & F1 & $\Delta$ F1  \\
    \midrule

   ConSeC$_{HEU}$  &      all &      80.4 &    64.3 &    71.5 & --  \\
   ConSeC$_{HEU}$  &      one &      71.6 &    56.4 &    63.1 & \textcolor{red}{~~-8.4}  \\
   ConSeC$_{HEU}$  &      no &      ~~0.0 &    ~~0.0 &    ~~0.0  & \textcolor{red}{-71.5}  \\
    \midrule
   Our Model  &  all  &  75.2 &    76.7 &    75.9  & --\\
   Our Model  &  one   & 70.4 &    73.1 &    71.7 & \textcolor{red}{~~-4.2}  \\
   Our Model  &  no   &  68.5 &    62.5 &    65.4 & \textcolor{red}{-10.5}  \\
    \bottomrule
    \end{tabular}
    \caption{WSL analysis on CG oracle.}
    \label{tab:wsd-ablation-results}
\end{table}

Table \ref{tab:wsl-results} reports the behavior of our model and the comparison systems. 
We note that all systems experience, in this unconstrained setting, a large drop in performance, with scores diverging markedly from those reported for WSD in Table \ref{tab:wsd-results} on \textbf{ALL}$_{FULL}$ (e.g., \num{82.5} vs \num{71.5} for ConSeC). 
All the same, our model behaves significantly better, attaining \num{75.9} (\num{5} point drop) and surpassing the best alternatives, i.e., ConSeC$_{HEU}$, by almost \num{5} points.
Interestingly, while our model does, in fact, lose to ConSeC$_{HEU}$ in terms of precision (\num{80.4} vs \num{75.2}), it displays large improvements in recall (\num{64.3} vs \num{76.7}). 

To better understand these results, we investigate how CD approaches fare in terms of span recall.\footnote{This refers to the accuracy of systems in identifying spans in ALL, irrespective of their associated senses.} What we find is that, while our model achieves \num{92.1}, the two baselines lag behind by a considerable margin, with $HEU$ and $SUP$ reaching \num{80.6} and \num{68.2}, respectively. 
Besides showing the effectiveness of our model on CD, this has several implications. 
First, tackling CD in a supervised fashion, in the current regime of WSL data, without taking into account the reference inventory, is not sufficient for good performance. 
Second, while $HEU$ is a better baseline, neglecting the semantics, i.e. choosing the longest span with at least one valid sense, is deleterious. 
Paired together, these findings suggest that hybrid schemes like ours, where both training examples and the inventory are jointly employed, represent a better alternative. 
Finally, while $HEU$ achieves nearly acceptable performance in English, the same does not hold for other languages where word inflection is a more complex phenomenon, and the mapping from word to senses lacks coverage compared to English\footnote{This refers to the resources available to the research community.}: for instance, testing span recall in Italian, using XL-WSD \cite{pasini-etal-xl-wsd-2021}, results in \num{70.7}, \num{10} points less than English.

\subsection{Word Sense Linking: Relaxing the Candidate Generation Oracle}
\label{sec:evaluation-wsd-cd-cg}

\paragraph{Setting and Comparison Systems} Here, we drop the R2 condition and examine different relaxations of R3.
That is, compared to the setting of the previous section, we assume the mapping from word to possible senses, which we used up to this point as our Candidate Generation oracle, to be either incomplete or absent. 
Such scenarios are realistic as they mimic low-resource language settings in which dedicated WordNet and the related word-to-sense mappings are incomplete.
We compare our model with ConSeC$_{HEU}$.


\paragraph{Data} 
To highlight the importance of having a complete mapping between senses and their lemmas, we analyze the performance of both systems in three different settings, that is, when, for each synset, we have at our disposal:
\begin{enumerate*}[label=\roman*)]
    \item all the lemmas with which it can be expressed,
    \item only its most frequent lemma, or
    \item no lemma at all.
\end{enumerate*}

In the first setting, where all possible lemmas are available, the system can utilize all the lexical variations for disambiguation. 
The second setting, which restricts the information to only the most frequent lemma, simulates a scenario with limited lexical resources, requiring the system to rely only on the most common representation of each sense. This may impact the system’s ability to accurately disambiguate less frequent word forms. The third setting, where no lemma information is provided, forces the system to rely solely on the definition, and this is the most challenging scenario. 

\paragraph{Our Model Behavior} Our model behavior is unchanged compared to the previous section. The only difference across the three settings regards the textual representation of each sense in $I$, which now consists of its definition postpended to \begin{enumerate*}[label=\roman*)]
    \item all its lemmas,
    \item only its most frequent lemma, or
    \item no lemma at all.
\end{enumerate*}

\paragraph{Results}
As evidenced by Table \ref{tab:wsd-ablation-results}, both systems perform worse when leveraging an incomplete mapping.
However, our model shows a more robust behavior in such a setting compared to ConSeC$_{HEU}$.
Indeed, as it performs the CG step independently from the mapping (with the retriever module), and takes advantage of the lemmas only to enhance sense representations, it is less reliant on this mapping.
In contrast, ConSeC$_{HEU}$ requires the sense mapping in order to retrieve the candidates for the WSD step.

More in detail, when limiting to only the main lemma for each sense, ConSeC$_{HEU}$ can only predict senses that are expressed by their primary lemma. 
As expected, this results in a decrease of \num{8.4} F1 points.
Our model, instead, is able to predict all senses, with a lower impact of \num{4.2} F1 points. Moving to the scenario where no lemmas are available, our system is still able to recognize and disambiguate spans, but its performance drops significantly by \num{10.5} F1 points. Yet, compared with ConSeC$_{HEU}$, this is still a huge improvement: indeed, ConSeC$_{HEU}$ is completely unable to perform the task as it lacks candidates.

Our objective is to evaluate our models' effectiveness in scenarios with limited CG oracles, typical of most mid- and low-resource languages in WSD. The results we report not only underline the robustness of our model but also highlight the inadequacy of modern WSD systems.

\section{Related Work}

Since this work is the first to introduce Word Sense Linking, unsurprisingly, there is no previous literature that covers it. However, what is surprising is the total absence, to the best of our knowledge, of studies that examine how recent state-of-the-art WSD systems scale to real-world scenarios. This is especially puzzling when we consider the growing interest that WSD has been receiving.
Indeed, thanks to the introduction of pretrained transformer-based language models \cite{devlin-etal-2019-bert, lewis-etal-2020-bart, he-deberta-2021}, this task has been witnessing renewed attention, with the research community focusing on challenging directions such as unseen  
prediction, cross-inventory generalization, and data efficiency, inter alia. Among the alternatives explored, the usage of definitions, which we also follow in this work, has proved to be particularly effective \cite{huang-etal-2019-glossbert, blevins-zettlemoyer-2020-moving, barba-etal-2021-esc}, achieving unprecedented performances and allowing sense representations to be disentangled from their occurrences in the training corpus.

Yet, in spite of this trend and its promising results, to the best of our knowledge, no assessment of how these models might actually be applied in general real-world scenarios has been made, thereby overlooking the relevance of Concept Detection and Candidate Generation. We suspect that this oversight is due mainly to two reasons:
\begin{enumerate*}[label=\roman*)]
    \item the performances on WSD benchmarks being too scarce, at least until recently, for any downstream application, and
    \item the research community assuming that adequate word sense mappings are always available for CG and that a simple heuristic approach could solve CD.
\end{enumerate*}
In our CD approach, we chose the greedy strategy, a decision supported by  \citet{MartnezRodrguez2020InformationEM}, who observed its widespread use in span identification within texts. However, our study clearly shows that is not a trivial task. Both heuristic and supervised techniques report definitely suboptimal behaviors. 

Finally, to provide some background on Candidate Generation, the task has generally been approached by striving to enumerate all the possible senses a word can assume. However, this is a prohibitively challenging endeavor, especially when wishing to scale across languages. While the research community has put forward a number of studies and mitigating strategies \cite{taghizadeh-faili-2016-automatic, al-tarouti-kalita-2016-enhancing, khodak-etal-2017-automated, neale-2018-survey}, the resulting resources are still incomplete. 

Arguably closest to our work, especially to our WSL model, is \citet{zhang-etal-2022-entqa}, who address Entity Linking by initially generating candidates, followed by Mention Detection (akin to our Concept Detection) and Entity Disambiguation (similar to our Word Sense Disambiguation). However, their reader architecture differs significantly from ours: they link one candidate at a time, whereas our model can simultaneously process all spans in an input sequence. A comparison with our model can be found in \autoref{sec:comparison_entqa}.

\section{Conclusion}

In this work, we challenge the assumptions behind Word Sense Disambiguation (WSD) and introduce a novel task called Word Sense Linking (WSL).
WSL requires a system to identify and disambiguate all the spans in an input text using only the information contained in a reference inventory, offering a scenario that is more aligned with practical downstream applications than the conventional WSD approach.
Along with the WSL formalization, we discuss a first comprehensive study in this direction, presenting a novel retrieved-reader architecture for the task, a complete and comprehensive benchmark for WSL systems, and an analysis of its performances and those of state-of-the-art WSD systems in multiple settings. 
Our findings highlight several important yet overlooked challenges that arise when scaling to unconstrained settings. 
In particular, natural expansions of WSD systems to WSL appear to be quite brittle, resulting in large performance drops. 
Conversely, our proposed architecture appears to be considerably more robust, achieving superior performances across all WSL settings.
Looking ahead, we plan to investigate the expansion of WSL to a multilingual setting and analyze the usage of WSL systems in downstream applications.

\section{Limitations}
This work has two inherent limitations: first, due to space constraints, we have deferred the evaluation of the model in a multilingual setting to future work. Potential challenges include the necessity for an extension of the sense inventories and the availability of training resources, which are requirements that go beyond the scope of the current study. Second, the lack of WSL-specific annotated data meant that we had to rely on datasets designed for Word Sense Disambiguation for training our models. Although these datasets offered valuable insights and exhibited promising results on our WSL-specific benchmarks, the prevalence of annotation gaps could hinder the performance of WSL systems. The effort to develop such datasets would be extensive, mirroring the significant undertaking required for our WSL-specific evaluation benchmark.

\section*{Acknowledgements}
\begin{center}
\noindent
    \begin{minipage}{0.1\linewidth}
        \begin{center}
            \includegraphics[scale=0.05]{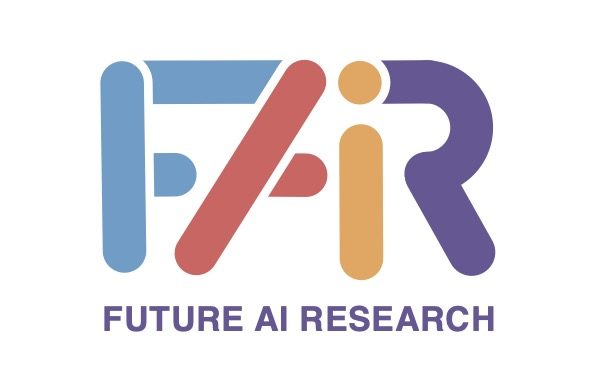}
        \end{center}
    \end{minipage}
    \hspace{0.01\linewidth}
    \begin{minipage}{0.70\linewidth}
         We gratefully acknowledge the support of the PNRR MUR project PE0000013-FAIR.
    \end{minipage}
    \hspace{0.01\linewidth}
    \begin{minipage}{0.1\linewidth}
        \begin{center}
            \includegraphics[scale=0.08]{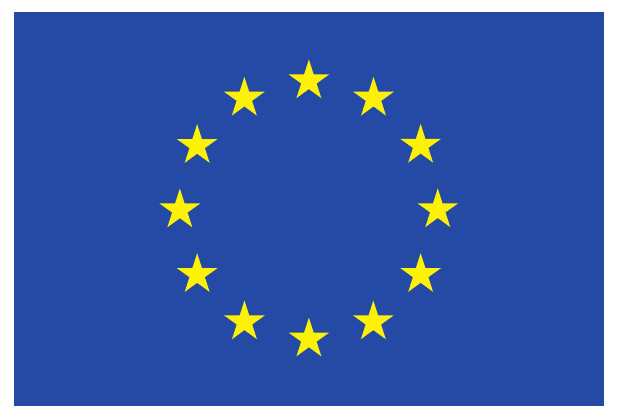}
        \end{center}
    \end{minipage}\\
\end{center}
\vspace{0.2cm}

\noindent We gratefully acknowledge the CREATIVE project (CRoss-modal understanding and gEnerATIon of Visual and tExtual content), which is funded by the MUR Progetti di Ricerca di Rilevante Interesse Nazionale programme (PRIN 2020). This work has been carried out while Andrei Stefan Bejgu was enrolled in the Italian National Doctorate on Artificial Intelligence run by Sapienza University of Rome. Edoardo Barba is fully funded by the PNRR MUR project PE0000013-FAIR.

\bibliography{anthology,custom}

\appendix

\section{Retriever Performances}
\label{sec:appendix}

In this section, we show the results of the Retriever module using the same hyperparameters for training as those described in Section \ref{sec:evaluation-model} and making just one change to the configuration showing the impact on the performances in terms of recall top-100 (R@100). We report the results of our experiments in Table \ref{tab:retriever-results}. The baseline retriever achieves a $96.5$ R@100, confirming our thesis that performing the Candidate Generation step is possible without knowing the spans a priori. Moreover, these results set a remarkable upper bound for the Reader module performances. The architecture of our baseline model is based on $bert-base-uncased$ \cite{devlin-etal-2019-bert} initialized with the weights from Sentence-Transformers \cite{reimers-gurevych-2019-sentence} and in particular using the weights of $E5_{base}$ \cite{wang2022text}. In this baseline setting, the textual representation of the senses of $I$ is the standard one, namely, that composed by the concatenation of all its available lemmas and its textual definition. We can see that initializing our weights from the generic $bert-base-uncased$ yield to $7.8$ points performance loss in recall shows that sentence-embedding pretraining is useful. Furthermore, using a more parameter-efficient architecture (33M, MiniLM-L6) compared to our reference one (109M, bert-base-uncased) still leads to competitive results ($92.5$). Finally, we can see that the standard setting where the senses of $I$ are represented with all the lemmas yields the best results; we gain $+4.0$ points over the textual representation composed by just one lemma and $+11.2$ points gain compared to the settings when lemmas are not used at all.

\begin{table}[t]
\center
\begin{tabular}{lc|c}
\toprule
Models & Params  & ALL R@100 ($\Delta$)  \\
\midrule
baseline &      109M &    96.5 \\
- bert-base-uncased &      109M &  88.7 (\textcolor{red}{-7.8})   \\
- $E5_{small}$ &      33M &    94.2 (\textcolor{red}{-2.3}) \\
- just main lemma &      109M &    92.5 (\textcolor{red}{-4.0}) \\
- no lemma &      109M &    85.3 (\textcolor{red}{-11.2}) \\
\bottomrule
\end{tabular}
\caption{Results in terms of the ablation study on the Retriever Module. Each row represents a change made to the baseline model and the corresponding impact on performance.}
\label{tab:retriever-results}
\end{table}

\section{Dataset Details}
\label{sec:appendix_ann}

\begin{figure}[h!]
    \centering
    \includegraphics{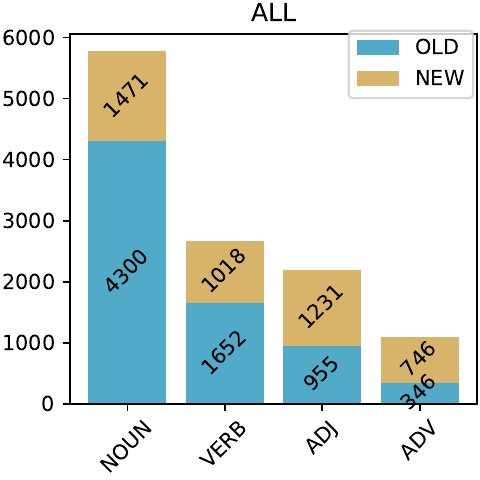}
    \caption{The counts of four POS categories within the 'ALL' dataset, which aggregates data across multiple sources. The 'OLD' (blue) bars represent the original annotation counts, while the 'NEW' (orange) bars indicate the increased counts following our extensive annotation process.}
    \label{fig:sem_stats_all}
\end{figure}

 We introduced a substantial addition of new instances across various datasets, achieving an overall 60\% increase as shown in \autoref{fig:sem_stats_all}.

\begin{figure*}[h!]
    \centering
    \includegraphics[width=1\textwidth,clip]{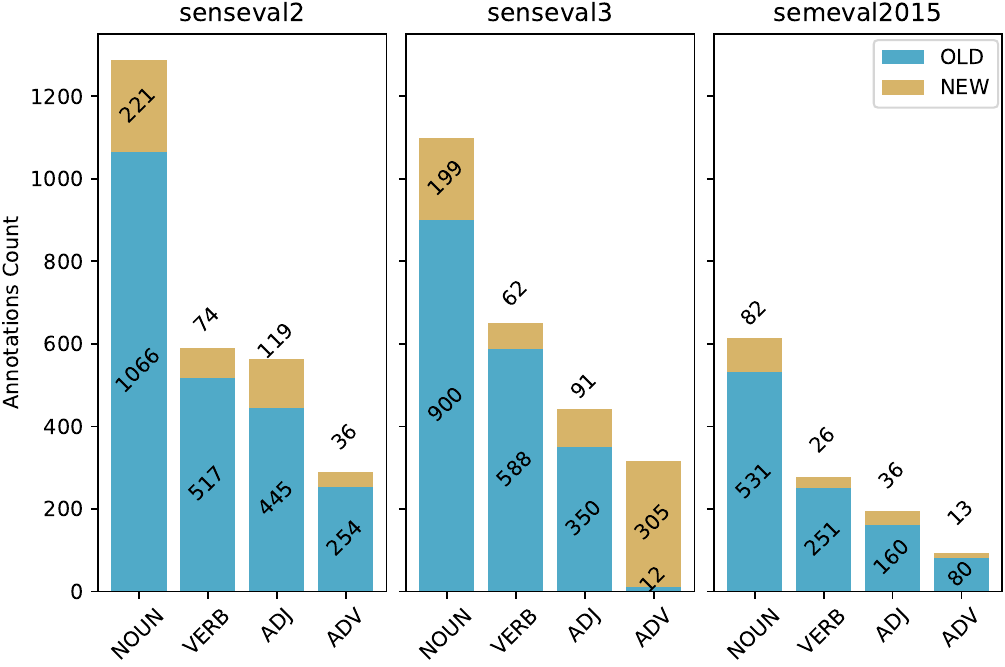}
    \caption{The counts of four POS categories (NOUN, VERB, ADJ, ADV) for three different datasets (Senseval2, Senseval3, Semeval2015). Each POS category is subdivided into 'OLD' (blue) and 'NEW' (orange) data points, indicating the frequency of each annotation before and after our comprehensive annotation process. }
    \label{fig:sem_stats}
\end{figure*}

\begin{figure}[h]
    \centering
    \includegraphics[width=0.9\textwidth,clip]{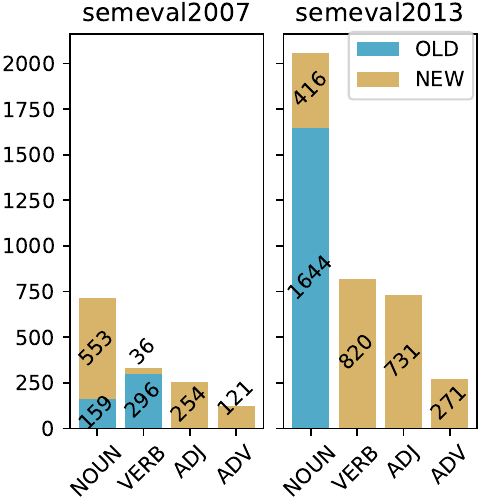}
    \caption{The count of POS categories for the Semeval2007 and Semeval2013 datasets. Notably, the original Semeval2007 dataset lacks annotations for ADJ and ADV categories, and Semeval2013 lacks annotations for VERB, ADJ, and ADV, as indicated by the absence of 'OLD' (blue) bars for these categories. The 'NEW' (orange) bars represent the counts post-annotation. }
    \label{fig:sem_stats_problematic}
\end{figure}

Before our annotation process, significant gaps were present in the POS tags across various datasets, with certain categories, such as verbs, adjectives, and adverbs in the semeval2013 dataset, and adjectives and adverbs in the semeval2007 dataset, being completely absent. This is evident from the missing data points in \autoref{fig:sem_stats_problematic}. Filling these gaps is crucial for constructing robust and comprehensive evaluation benchmarks. Incomplete datasets can lead to evaluations that fail to measure the true capabilities of language processing systems in real-world scenarios. Our annotation efforts were, therefore, critical in ensuring that all POS categories were fully represented, thereby enhancing the validity and reliability of subsequent system evaluations. Moreover, we preserved the distribution across POS tags as shown in \autoref{fig:sem_stats}.

\begin{table*}[t]
    \center
    \small
    \begin{tabular}{cl|llll}
    \toprule
    & Dataset & Sentences & Tokens & Instances & New Instances \\
    \midrule
    \parbox[t]{2mm}{\multirow{2}{*}{\rotatebox[origin=c]{90}{\textit{Train}}}}
    & SemCor &      37176 &      820410 & 226036 & ~~~- \\
    & SemCor$_C$ &      37176 &      820410  & 359763 & ~~~-  \\
    \midrule
    \parbox[t]{2mm}{\multirow{6}{*}{\rotatebox[origin=c]{90}{\textit{Evaluation}}}}
    & semeval2007 &      135 &      3219 & ~~455 & ~~941 (\textcolor{teal}{+206\%}) \\
    & semeval2013 &      306 &      8533  & 1644 & 2194 (\textcolor{teal}{+133\%}) \\
    & semeval2015 &      138 &      2643  & 1022 & ~~157 (\textcolor{teal}{+15\%}) \\
    & senseval2 &      242 &      5829  & 2282 & ~~444 (\textcolor{teal}{+19\%}) \\
    & senseval3 &      352 &      5640  & 1850 & ~~634 (\textcolor{teal}{+34\%}) \\
    & all &      1173 &      25864 & 7253 & 4370 (\textcolor{teal}{+60\%}) \\
    \bottomrule
    \end{tabular}
    \caption{Statistics for training and evaluation corpora. The columns represent the number of sentences, the total number of tokens, the number of annotated terms, and the number of newly annotated instances added in each dataset.}
    \label{tab:wsd-dataset-stats}
\end{table*}

\begin{table}[t]
    \center
    \small
    \begin{tabular}{c|l|l}
\toprule
     Models  & SemCor  & SemCor$_C$  \\
    \midrule
        BEM   & 79.0 & 78.8 (\textcolor{red}{-0.2}) \\

    ESCHER                      &    80.7  &        80.3 (\textcolor{red}{-0.4})  \\
     ConSeC  &        82.0  &     81.2 (\textcolor{red}{-0.8}) \\
    \bottomrule
    \end{tabular}
    \caption{WSD F1 score results on the SemCor$_C$ the dataset containing the silver annotations annotations from ConSec$_{HEU}$.}
    \label{tab:wsd-semcor_c-all}
\end{table}

\begin{table}[t]
    \center
    \small
    \begin{tabular}{lcccccc}
    \toprule
    \multicolumn{1}{c}{} & \multicolumn{3}{c}{\textbf{SE07}} & \multicolumn{3}{c}{\textbf{ALL$_{FULL}$}} \\ 
    \cmidrule(lr){2-4} \cmidrule(lr){5-7}
    Models & P & R & F1 & P & R & F1 \\
    \midrule
    ConSeC$_{HEU}$ & \textbf{76.7} & 55.4 & 64.3 & \textbf{80.4} &   64.3 & 71.5 \\
    EntQA &  75.1 & 64.7 & 69.5 & 78.4 & 66.5 & 72.0\\

    Our Model & 73.8 & \textbf{74.9} & \textbf{74.4} & 75.2 & \textbf{76.7} & \textbf{75.9} \\
    \bottomrule
    \end{tabular}
    \caption{Our model comparison with EntQA in the WSL task tested on ALL$_{FULL}$ dataset.}
    \label{tab:wsl-results-entqa}
\end{table}

\section{Evaluation of SemCor with ConSeC$_{HEU}$ Annotations}
\label{sec:app_semcor}
Given the known limitations of SemCor due to its incomplete annotations, we have devised a mitigation strategy. By employing ConSeC$_{HEU}$, we aimed to identify and annotate all missing spans within SemCor, thereby creating an enhanced version, SemCor$_C$. This process, as detailed in \autoref{tab:wsd-dataset-stats}, resulted in the addition of 133,727 new annotations to the existing 226,036 in SemCor. Subsequently, we assessed the performance of leading Word Sense Disambiguation systems ESC, BEM, and ConSeC using this enriched dataset. However, as indicated in \autoref{tab:wsd-semcor_c-all}, the enhancements in SemCor$_C$ did not necessarily translate into improved performance of the WSD models; with BEM achieving an F1 score of 79.0, ESCHER at 80.7, and ConSeC at 82.0. Upon transitioning to the SemCor$_C$ dataset, a noticeable decline in performance is observed for each model: BEM experiences a slight reduction to 78.8 (a 0.2 decrease), ESCHER to 80.3 (a 0.4 decrease), and ConSeC exhibits a more significant drop to 81.2 (a 0.8 decrease). We argue that this outcome originates from the 'silver' quality of the newly added annotations. 

However, despite the quality of the new annotations in SemCor$_C$ not matching the 'gold' standard of the original dataset, it plays an important role in our WSL setting. This integration improves the model's capacity for the production of accurate representation and identification of spans in the text.

\section{Comparison with EntQA}
\label{sec:comparison_entqa}

In this section, we present a detailed comparison between EntQA \cite{zhang-etal-2022-entqa}, and our model in the context of a WSL unconstrained environment. While EntQA represents a significant benchmark in the field, particularly for tasks similar to our WSL model, such as Entity Linking through the pipeline: candidate generation, Mention Detection (comparable to our Concept Detection), and Entity Disambiguation (analogous to our Word Sense Disambiguation), the two architectures diverge particularly in the reader part.  Unlike EntQA, which processes candidates sequentially, our model can simultaneously handle all spans within an input sequence. We used the same Retriever for both reader models. In particular, we re-implemented the EntQA reader model. 

The performance outcomes, as shown in \autoref{tab:wsl-results-entqa}, show that our model outperforms EntQA. Our model obtained an F1 score of 75.9 compared to EntQA's 72.0. This superiority in performance underscores the effectiveness of our model, especially in terms of recall and overall F1 score.  
We argue that, in our model, contextualizing all the candidates of a sentence together plays a crucial role in the performance gain.
Moreover, our model not only outperforms EntQA but it is also more efficient. By processing all candidates together, our model significantly reduces processing times, in contrast to EntQA's sequential candidate method. This efficiency is quantitatively evident as our system processes the ALL$_{FULL}$ dataset in merely 17 seconds, a substantial improvement over the 63 seconds required by EntQA. To perform the comparison, we evaluated the same machine using an RTX 3090 for both experiments. This speed-up not only demonstrates our model's enhanced performance in terms of speed but also reinforces its practicality for integration into downstream tasks, further establishing our method's advantage in the field of WSL unconstrained settings.

\section{Qualitative analysis}
\label{sec:appendix_qual}
In this section, we present a qualitative analysis of the model's output. We identified instances that highlight mismatches and inherent data discrepancies rather than direct errors in model processing. These cases highlight the complexities of matching model interpretations with established lexical databases such as WordNet. Some examples include:

\begin{itemize}
    \item \textbf{Lexical Variants and Inventory Gaps:} The model can identify terms that are absent in the sense inventory, highlighting a gap between model recognition capabilities and standardized lexical entries. Some examples are shown in Table~\ref{tab:qual_gap}. For instance, in the sentence \emph{``training and development of ageing workers in both the work place and the community,''} the model accurately identifies the span \texttt{[work place]} and annotates it with the sense \emph{[a place where work is done]}. This instance reveals a mismatch due to the lexical variant ``work place'' not being directly mapped to its standard form ``workplace'' in WordNet. 

    \item \textbf{Named Entities and Sense Attribution:}  The model tends to abstract named entities into broader conceptual categories, as shown in Table~\ref{tab:qual_ent}. For instance, in the sentence the sentence \emph{``Trouble is following hard on the heels of the uproar around Josef Ackermann, CEO of Deutsche Bank,''} the model categorizes \texttt{[Josef Ackermann]} as \emph{[the corporate executive responsible for the operations of the firm;]}. Although WordNet contains some named entities, specific ones like ``Josef Ackermann'' may not be explicitly available in the inventory. Moreover, in the Semcor training samples, often named entities are annotated with generic synsets like \emph{person} or \emph{location}, illustrating the challenge of capturing the full specificity of named entities within existing sense inventories.
\end{itemize}

These examples highlight WSL's challenges with lexical variations and named entity interpretation, emphasizing the importance of refining sense inventories and training methods for better alignment with lexical standards.

\begin{table*}[h!]
\centering
\begin{tabular}{|m{7cm}|m{7cm}|}
\hline
\textbf{Example Text} & \textbf{WSL disambiguation} \\ 
\hline
Training and development of
ageing workers in both the  \textcolor{red}{work place} and the community. & a place where work is done \\
\hline
In the amount USD 45 billion ( nearly   \textcolor{red}{EUR} 30 billion ) in one go . & the basic monetary unit of most members of the European Union \\
\hline
Auditors found \textcolor{red}{crookery} the first day on the job. & verbal misrepresentation intended to take advantage of you in some way \\
\hline
Played on the 23rd of November against \textcolor{red}{Ajax} in European Champions League& - any number of entities (members) considered as a unit; \\ &- an active diversion requiring physical exertion and competition  \\
\hline
Ctrl Q Quit \textcolor{red}{Shuts} the program. & cease to operate or cause to cease operating \\
\hline
\end{tabular}
\caption{This table showcases examples of model's disambiguation capabilities and lexical recognition gaps, showing specific instances where the model accurately identifies and annotates lexical variants not directly mapped in standard sense inventories }
\label{tab:qual_gap}

\end{table*}

\begin{table*}[h!]
\centering
\begin{tabular}{|m{7cm}|m{7cm}|}
\hline
\textbf{Example Text} & \textbf{WSL disambiguation} \\ 
\hline
Trouble is following hard on the heels of the uproar around \textcolor{red}{Josef Ackermann}, CEO of Deutsche Bank. & the corporate executive responsible for the operations of the firm; \\
\hline
In his program, \textcolor{red}{François Hollande} confines himself to banalities. & a human being \\
\hline
The \textcolor{red}{World Labor Organisation} estimates that for example in Germany.. & an international alliance involving many different countries \\
\hline
Friendly game today, at 3:05 pm at the \textcolor{red}{National Stadium} in San Jose.& location, a point or extent in space \\
\hline
The two justices have been attending \textcolor{red}{Federalist Society} events for years. & any number of entities (members) considered as a unit \\
\hline
\end{tabular}
\caption{This table showcases examples of how the model abstracts named entities into broader conceptual categories. Each row shows the model's disambiguation of specific named entities.}
    \label{tab:qual_ent}

\end{table*}

\textcolor{white}{-}  
\newpage 
\textcolor{white}{-}


\end{document}